\documentclass[final,5p,times,twocolumn,authoryear]{elsarticle}

\usepackage{amssymb}

\usepackage{natbib}
\usepackage{hyperref}
\usepackage{framed,multirow}

\usepackage{amsmath,amssymb,amsfonts}
\usepackage{algorithmic}
\usepackage{graphicx}
\usepackage{textcomp}

\usepackage{makecell}
\usepackage{bm}
\usepackage{array}
\usepackage{threeparttable}
\usepackage{color}
\definecolor{R}{rgb}{1,0,0}
\usepackage{booktabs,diagbox,float}

\usepackage{algorithm}
\usepackage{algorithmic}
\journal{ }

\usepackage{color}

\begin{document}

\begin{frontmatter}



\title{LDM-Morph: Latent diffusion model guided deformable image registration}

\author[]{Jiong Wu}
\author[]{Kuang Gong\corref{cor1}}
\cortext[cor1]{
Corresponding author: Kuang Gong (kgong@bme.ufl.edu)}
\address[]{J. Crayton Pruitt Family Department of Biomedical Engineering, University of Florida, Gainesville, FL, 32611, USA}


\begin{abstract}
	Deformable image registration plays an essential role in various medical image tasks. Existing deep learning-based deformable registration frameworks primarily utilize convolutional neural networks (CNNs) or Transformers to learn features to predict the deformations. However, the lack of semantic information in the learned features limits the registration performance. Furthermore, the similarity metric of the loss function is often evaluated only in the pixel space, which ignores the matching of high-level anatomical features and can lead to deformation folding. To address these issues, in this work, we proposed LDM-Morph, an unsupervised deformable registration algorithm for medical image registration. LDM-Morph integrated features extracted from the latent diffusion model (LDM) to enrich the semantic information. Additionally, a latent and global feature-based cross-attention module (LGCA) was designed to enhance the interaction of semantic information from LDM and global information from multi-head self-attention operations. Finally, a hierarchical metric was proposed to evaluate the similarity of image pairs in both the original pixel space and latent-feature space, enhancing topology preservation while improving registration accuracy. Extensive experiments on four public 2D cardiac image datasets show that the proposed LDM-Morph framework outperformed existing state-of-the-art CNNs- and Transformers-based registration methods regarding accuracy and topology preservation with comparable computational efficiency. Our code is publicly available at: \url{https://github.com/wujiong-hub/LDM-Morph}.
\end{abstract}


\end{frontmatter}


\section{Introduction}

Deformable image registration is a fundamental step in various clinical tasks. It aligns moving images with fixed ones through dense nonlinear spatial mapping, accommodating anatomical variations across inter-patients or different stages of disease in intra-patients. This precise alignment supports automated analysis, which is crucial for preoperative planning~\citep{torchia2024deformable}, intraoperative information fusion~\citep{gopalakrishnan2024intraoperative}, disease diagnosis~\citep{cui2023deep}, and patient follow-ups~\citep{alam2018medical}. 

Over the past few years, the rapid advancements in deep learning, especially the exceptional performance of convolutional neural networks (CNNs) in image classification~\citep{rawat2017deep}, semantic segmentation~\citep{dong2024head}, and object detection~\citep{litjens2017survey}, have led to their application in medical image registration~\citep{hering2022learn2reg}. This shift has revolutionized the registration field, transitioning from traditional optimization-based methods to learning-based paradigms. Unlike traditional registration algorithms, deep learning-based registration (DLR) methods leverage feature learning, driven by predefined loss functions, to estimate the anatomical correspondence inside the image pair~\citep{wu2024diffeomorphic}. As a result, the registration performance of such a feature learning-based framework is highly dependent on the feature-extraction capability of the neural networks.

One issue of CNNs is that the localized receptive fields limit their ability to learn long-range dependencies, which are crucial for image registration as large deformations between images might exist, especially for the datasets with pathological changes~\citep{chen2022transmorph}. Recently, Transformers have been successfully applied to medical image analysis tasks, such as disease prediction~\citep{huo2024hifuse} and region of interest (ROIs) segmentation~\citep{tang2022self}, and achieved better performance than CNNs due to its powerful multi-head self-attention mechanism~\citep{chen2021crossvit} in capturing long-range dependencies. The Swin Transformer-based registration architectures have become mainstream solutions for medical image registration with large deformations~\citep{chen2022transmorph,chenz2023transmatch}. 

Despite the superior performance of Transformer-based frameworks over CNN-based methods in most registration tasks, the features learned directly from these networks using predefined loss functions may not fully capture deformation-related information, and further improvement is still needed. Additionally, existing DLR methods have mainly been applied to magnetic resonance imaging (MRI) and  Computed Tomography (CT) images. Applying these methods to other modalities, such as cardiac ultrasound, remains challenging due to the lower contrast and higher noise levels in ultrasound images compared to MRI and CT. Finally, dissimilarity metrics in existing DLR methods are often calculated solely in the original pixel space, neglecting high-level semantic information, which can lead to anatomical topology mismatching and further hinder registration performance.

To overcome these challenges, in this work, we proposed LDM-Morph, a dual-stream cross-learning framework designed for unsupervised deformable medical image registration, leveraging the powerful feature-extraction capability of latent diffusion model (LDM)~\citep{rombach2022high}. Compared with existing deformable image registration methods, including conventional and DLR methods, the proposed LDM-guided framework eliminated the need for time-consuming iterative optimization to obtain optimal deformations, effectively captured diverse image features for more precise registration, and excelled in aligning high-level anatomical features with enhanced accuracy. 
 
To be specific, the major contributions of this work are listed as follows:
\begin{itemize}
\item 
In this proposed LDM-Morph framework, LDM was leveraged as a feature extractor to obtain high-level semantic information, which helped better align the image pairs in the feature space for image registration tasks.

\item 
Inside the dual-stream encoder proposed, a latent and global feature-based cross-attention (LGCA) module was designed to capture and integrate the latent and global features from the LDM and the multi-head self-attention modules. 

\item
A hierarchical similarity metric was introduced within the framework to match images at both the pixel level and the semantic level. This straightforward yet effective penalty metric enhanced registration accuracy while reducing deformation folding.

\end{itemize}

The remainder of this paper is organized as follows. Sec.~\ref{section2} introduces the related work in medical image registration. Sec.~\ref{section3} describes details of the proposed method. Sec.~\ref{section4} outlines the experimental setup while Sec.~\ref{section5} presents the results. Section~\ref{section6} presents a detailed discussion of this work and Sec.~\ref{section7} concludes the paper.

\section{Related work}
\label{section2}
\subsection{Conventional deformable image registration algorithms}
Conventional deformable image registration methods were built on physics systems, and optimal deformations were calculated by minimizing predefined objective functions. For example, Beg et al.~\citep{beg2005computing} proposed the Large Deformation Diffeomorphic Metric Mapping (LDDMM) algorithm grounded in fluid mechanics. LDDMM defined an energy function that incorporated a Laplacian differential operator applied to a series of time-dependent velocity fields, ensuring that the resulting deformation was a diffeomorphism~\citep{zhang2015finite, wu2020large}. Building on LDDMM, the Symmetric Normalization (SyN) algorithm introduced a symmetric diffeomorphic registration framework. SyN employed normalized cross-correlation (NCC) as the matching cost function, guiding the moving and fixed images to warp towards their midpoint, thereby enhancing accuracy and preserving anatomical topology~\citep{avants2008symmetric}. Similarly, Thirion et al.~\citep{vercauteren2009diffeomorphic} addressed the registration problem using an optical flow model, formulating deformation estimation as a solution to the optical flow equation. While these traditional algorithms excelled in preserving topology, they relied on gradient descent and other optimization techniques that must navigate high-dimensional parameter spaces. This process required numerous iterations to manage the high degrees of freedom, leading to significant computational demands and extended processing times.

\subsection{Learning-based deformable registration methods}
Unlike conventional methods, which relied heavily on predefined models and optimization techniques, learning-based approaches used data-driven mechanisms to train neural networks to handle complex deformations between image pairs~\citep{kuppala2020overview}. In recent years, numerous supervised and unsupervised approaches have been proposed for medical image registration, with CNNs commonly adopted as the base networks~\citep{balakrishnan2019voxelmorph,de2019deep}. Unsupervised learning methods, such as VoxelMorph, trained CNNs directly using a predefined loss function that mirrored the objective function of traditional algorithms~\citep{balakrishnan2019voxelmorph}. Additionally, several studies have proposed multi-resolution registration frameworks~\citep{mok2020large,wu2024diffeomorphic}, coarse-to-fine strategies~\citep{meng2024correlation}, and multi-stage cascading registration architectures~\citep{xu20234d} to further improve registration performance.

Besides CNNs, Transformer-based architectures have also been explored for medical image registration. Chen et al.~\citep{chen2022transmorph} proposed a novel registration architecture, TransMorph, which replaced the encoder layers of U-Net with Swin-Transform blocks~\citep{liu2021swin}. Due to the enhanced feature learning enabled by the multi-head self-attention mechanism, TransMorph achieved higher accuracy than CNN-based methods. Ma et al.~\citep{ma2022symmetric} employed a Transformer-based block in the encoder and decoder to model the long-range spatial cross-image relevance for deformable image registration. Focusing on 2D image registration, Ding et al.~\citep{ding2024c2fresmorph} proposed a two-stage framework using CNN and the Transformer to capture local and global information sequentially. Regarding the cross-attention mechanism in Transformer, Chen et al.~\citep{chenz2023transmatch} introduced an end-to-end Swin Transformer-based framework, where cross-attention modules were proposed to match multilevel features and extract relevant information between moving and fixed images. To effectively capture short- and long-range flow features across multiple scales, Ghahremani et al.~\citep{ghahremani2024h} designed a hierarchical Vision Transformer with self- and cross-attention modules in a pyramid-like framework, enhancing robustness in deformable image registration. 

Inspired by these prior arts, in this work, we designed a novel cross-attention module, as described in Sec.~\ref{Dual}, to integrate latent features from LDM with global features from the multi-head self-attention mechanism.


\begin{figure*}[!t]
\includegraphics[width=16cm]{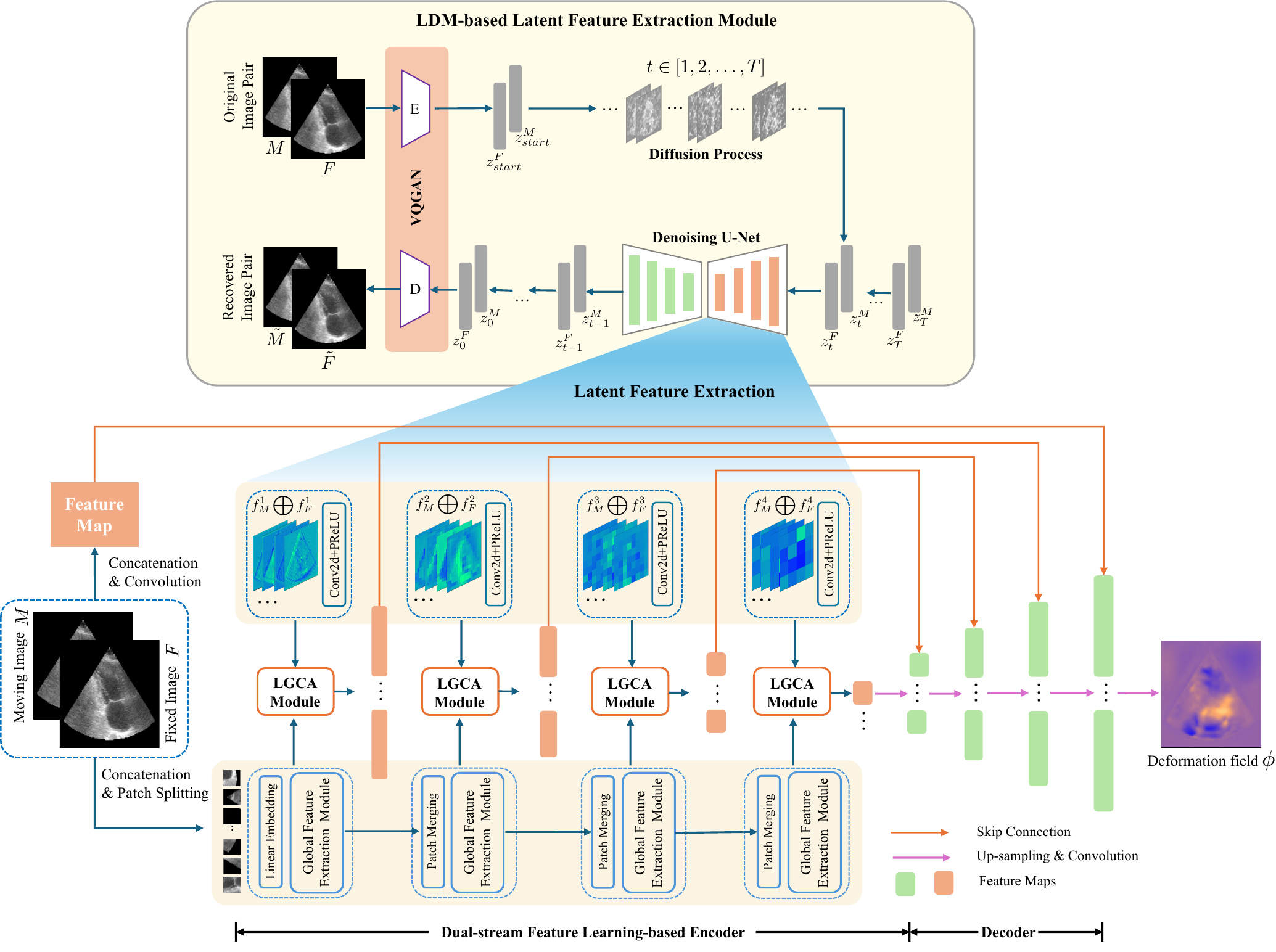}
\centering
\caption{Overview of the proposed deformable registration framework (LDM-Morph). It comprised three main components: a latent diffusion model (LDM)-based latent feature extraction (LDM-FE) module (shown on the top, detailed in Sec.~\ref{LDM}), a dual-stream feature learning-based encoder (detailed in Sec.~\ref{Dual}), and a decoder. The LDM was pre-trained. Afterward, the features extracted through the encoder of the U-NET in LDM were utilized in the dual-stream feature learning-based encoder. The estimated deformation field $\phi$ was outputted with the same resolution as the original images.} \label{fig1}
\end{figure*}

\subsection{Diffusion model-based registration approaches}
Score function-based diffusion models (DMs), particularly the denoising diffusion probabilistic model (DDPM)~\citep{ho2020denoising}, have demonstrated exceptional effectiveness in various medical image generation tasks~\citep{jang2023taupetgen,gong2024pet,jiang2024fast}. For image registration, Kim et al.~\citep{kim2022diffusemorph} proposed DiffusionMorph, adapting DDPM to generate deformation fields. Zhang et al.~\citep{zhang2023adaptive} used DDPM to model error residuals between image pairs, adaptively adjusting the smoothness constraint during optimization. Leveraging the extracted features from DM, Qin and Li~\citep{qin2023fsdiffreg} proposed a DM-guided deformable image registration framework for cardiac imaging.

Beyond DDPM, LDM~\citep{rombach2022high} has also gained attention for its memory and computational efficiency regarding image generation. LDM conducted the diffusion process in a latent space wherein higher-level semantic features were captured. In this work, as described in Sec.~\ref{LDM}, we embedded the semantic information extracted from LDM into the registration network to better align the image pairs. Accordingly, as described in Sec.~\ref{LossFunction}, a hierarchical similarity metric was utilized as part of the loss function to quantify the image dissimilarity from both original and latent space.


\section{Method}
\label{section3}
\subsection{Overall architecture}
Let $M$ and $F$ denote a moving and a fixed image defined in the background space $\Omega\in{R^2}$ with the size of $H\times{W}$, respectively. Our goal is to find a deformation field $\phi:\Omega\rightarrow\Omega$ such that the transformed moving image $M\circ{\phi}^{-1}$ can be well warped to the fixed image $F$. In this work, a novel network was designed to directly predict the deformation field $\phi=\Phi_\theta(M, F)$, where $\Phi_\theta$ represented the network with the learnable parameters $\theta$. The main components of the proposed network $\Phi_\theta$ consisted of an LDM-based feature extraction (LDM-FE) module, a dual-stream feature learning-based encoder, and an upsampling and convolution-based decoder. The diagram of the proposed framework is shown in Fig. \ref{fig1}. 

\subsection{LDM-based latent feature extraction module}
\label{LDM}
%
One key component of the proposed framework was the LDM-FE module, which acted as a feature extractor as shown in the top panel of Fig. \ref{fig1}.  For this LDM-FE module, an LDM was pretrained based on the image pairs as described below.

Within the LDM framework, given an arbitrary image $\mathcal{I}\in\Omega$, an encoder $E$  first transformed  $\mathcal{I}$ from the original pixel space to the latent space, yielding the latent variable $z_{start}^\mathcal{I}=E(\mathcal{I})$. Afterward, a diffusion process was conducted by gradually adding  Gaussian noise $\epsilon$ to the latent variable $z_{start}^\mathcal{I}$. This process could be viewed as a Markov transformation, in which the generated latent variable $z_t^\mathcal{I}$ at the $t$-th step was defined as
\begin{equation}
q(z_t^\mathcal{I}|z_{t-1}^\mathcal{I})=\mathcal{N}(z_t^\mathcal{I};\sqrt{1-\beta_t}z_{t-1}^\mathcal{I},\beta_tI),
\end{equation}
where $0\leq\beta_t\leq1$ was the variance of the Gaussian noise added. By leveraging the additive properties of the normal distribution, the distribution of $z_t^\mathcal{I}$ can be  obtained given $z_{start}^\mathcal{I}$, i.e.,
\begin{equation}
q(z_t^\mathcal{I}|z_{start}^\mathcal{I})=\mathcal{N}(z_t^\mathcal{I};\sqrt{\alpha_t}z_{start}^\mathcal{I},(1-\alpha_t)I),
\end{equation}
where $\alpha_t=\prod_{s=1}^t(1-\beta_s)$. Accordingly, given $\epsilon \sim\mathcal{N}(0,I)$,
\begin{equation}
z_{t}^\mathcal{I}=\sqrt{\alpha_t}z_{start}^\mathcal{I}+\sqrt{1-\alpha_t}\epsilon.
\end{equation} The reverse diffusion process was to learn the data distribution $p(z_{start}^\mathcal{I})$ via progressive denoising starting from $z_T^\mathcal{I}\sim\mathcal{N}(0,I)$. It could be considered as a reverse process of the Marhov Chain with the following form
\begin{equation}
p_{\omega}(z_{t-1}^\mathcal{I}|z_{t}^\mathcal{I})=\mathcal{N}(z_{t-1}^\mathcal{I};\mu_{\omega}(z_t^\mathcal{I},t),\sigma_t^2I),
\end{equation}
where $\sigma^2_t$ was the variance and $\mu_{\omega}(z_t^\mathcal{I},t)$ was a learned mean defined as
\begin{equation}
\mu_{\omega}(z_t^\mathcal{I},t)=\frac{1}{\sqrt{1-\beta_t}}\left(z_t^\mathcal{I}-\frac{\beta_t}{\sqrt{1-\alpha_t}}\epsilon_{\omega}(z_t^\mathcal{I},t)\right),
\end{equation}
where $\epsilon_{\omega}$ was a parameterized model through a U-Net in our implementation and trained by  the following loss function
\begin{equation}
\mathcal{L}_{LDM}=\mathbb{E}_{E(\mathcal{I}),\epsilon\thicksim\mathcal{N}(0,1),t}\left[\|\epsilon-\epsilon_\omega(z_t^\mathcal{I},t)\|_2^2\right].
\end{equation} 

During the reverse diffusion process, U-Net continuously encoded the features of $z_t^\mathcal{I}$ with the corresponding time embedding to predict the noise added in the $t$-th step. Since U-Net was used to estimate the noise at any step $t\in\{1,2,\dots, T\}$ with $T=1000$, such an extensive range of $t$ ensured that the latent features of $z_{start}^\mathcal{I}$ could be sufficiently captured by the U-Net~\citep{pnvr2023ld}.  Additionally, the learned latent features were rich in spatial information, which could benefit image alignment~\citep{kim2022diffusemorph}.  Therefore, we directly extracted the latent features from the U-Net encoder and embedded them into the upper stream of the dual-stream feature learning-based encoder.

In summary, after training the LDM,  the parameters of the encoder $E(\cdot)$ and U-Net were frozen.  The moving and fixed images, $M$ and $F$, were then encoded into the latent space, followed by the diffusion processes using DDIM inversion with $t$ steps to yield $z_t^M$ and $z_t^F$, respectively.  With inputs $(z_t^M, t)$ and $(z_t^F, t)$, a set of latent features $\{f_M^i\bigoplus{f_F}^i|i=1,2,\dots,n\}$ were extracted from the $n$ convolutional layers of the encoder component of the U-Net, where $n$ was set to 4 in our study. 

\subsection{Dual-stream feature learning-based encoder}
\label{Dual}
A dual-stream feature learning-based encoder consisted of two streams: the upper stream processed the latent features, while the lower one processed the original image pair. For the upper stream, three groups of latent features from the LDM-FE module were fed into three convolutional layers to ensure that the number of features matched that of the lower stream at each level. Regarding the lower stream, a global feature extraction module was introduced based on the multi-head self-attention mechanism. 
To fully exploit the relationship between latent and global features, a latent and global feature-based cross-attention (LGCA) module was proposed, which operated simultaneously on the outputs of both the upper and lower streams at each resolution level.


\begin{figure}[!t]
\includegraphics[width=5.0cm]{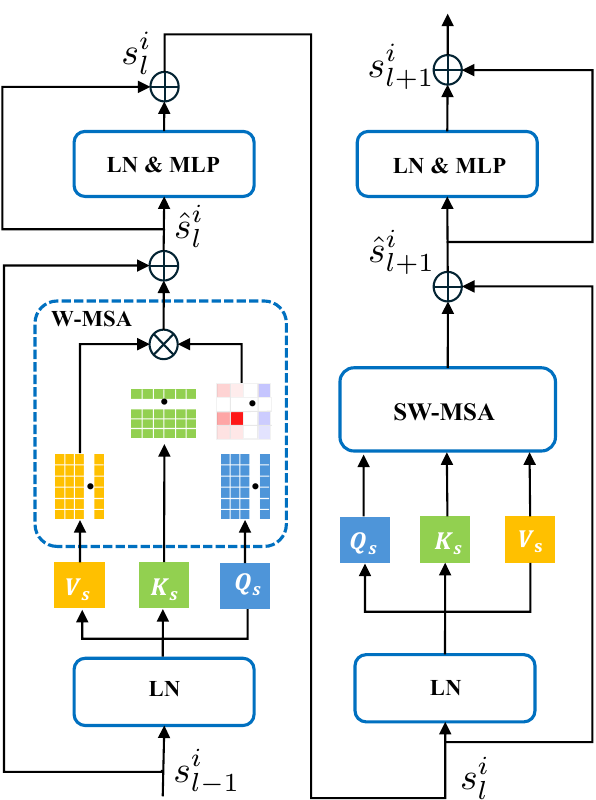}
\centering
\caption{The architecture of the global feature extraction module.} \label{fig2}
\end{figure}

\subsubsection{Global feature extraction module}
The global feature extraction module in our proposed LDM-Morph framework resembled the self-attention mechanism of Swin-Transformer~\citep{liu2021swin}.  As shown in Fig. \ref{fig2}, it consisted of two different attention strategies, including a window-based multi-head self-attention (W-MSA) and shift-window-based multi-head self-attention (SW-MSA). Feature maps from the previous operation were evenly divided into $W_x\times{W_y}$ non-overlapping windows, each containing $P_x\times{P_y}$ non-overlapping patches. W-MSA treated patches as tokens and calculated the self-attention inside each window. SW-MSA calculated the self-attention within adjacent windows using the Swin-Transformer's shift-window strategy. Specifically, each window was first shifted $\left(\lfloor\frac{W_x}{2}\rfloor,\lfloor\frac{W_y}{2}\rfloor\right)$ and then self-attention was computed inside each shifted window. Using window-based attention, the global feature extraction module conducted the calculations as follows,

\begin{equation}
\label{eq7}
\begin{aligned}
	\hat{\mathbf{s}}^i_l& =\text{W-MSA}(\mathrm{LN}(\mathbf{s}^i_{l-1}))+\mathbf{s}^i_{l-1}, \\
	\mathbf{s}^i_{l}& =\mathrm{MLP}(\mathrm{LN}(\mathbf{\hat{s}^i}_l))+\mathbf{\hat{s}}^i_l, \\
	\mathbf{\hat{s}}^i_{l+1}& =\text{SW-MSA}(\text{LN}(\mathbf{s}^i_l))+\mathbf{s}^i_l, \\
	\mathbf{s}^i_{l+1}& =\mathrm{MLP}(\mathrm{LN}(\mathbf{\hat{s}}^i_{l+1}))+\mathbf{\hat{s}}^i_{l+1}. 
\end{aligned}
\end{equation}
Here, $\text{W-MSA}(\cdot)$ and $\text{SW-MSA}(\cdot)$ represented regular- and shift-window partitioning configuration-based multi-head self-attention, respectively; $\text{LN}(\cdot)$ denoted the layer normalization, and $\text{MLP}(\cdot)$ denoted the multilayer perceptron module. $\mathbf{s}^i_{l-1}$ represented the input of the $i$-th global feature extraction module, while $\mathbf{s}^i_{l}$ and $\mathbf{s}^i_{l+1}$ represented the output features of  $\text{W-MSA}$ and $\text{SW-MSA}$. The self-attention was computed as
\begin{equation}
\textbf{Self-Attention}(Q_s,K_s,V_s)=\mathrm{softmax}(\frac{Q_sK_s^\top}{\sqrt{d}})V_s.
\end{equation}
Here $Q_s,K_s,V_s\in\mathbb{R}^{P_xP_yd}$ denoted the \textit{query}, \textit{key}, \textit{value} matrics, respectively. $d$ denoted the dimension of \textit{query} and \textit{key}, and $P_xP_y$ was the number of tokens in the 2D window. 

\subsubsection{LGCA module}
\begin{figure}[!t]
\includegraphics[width=6cm]{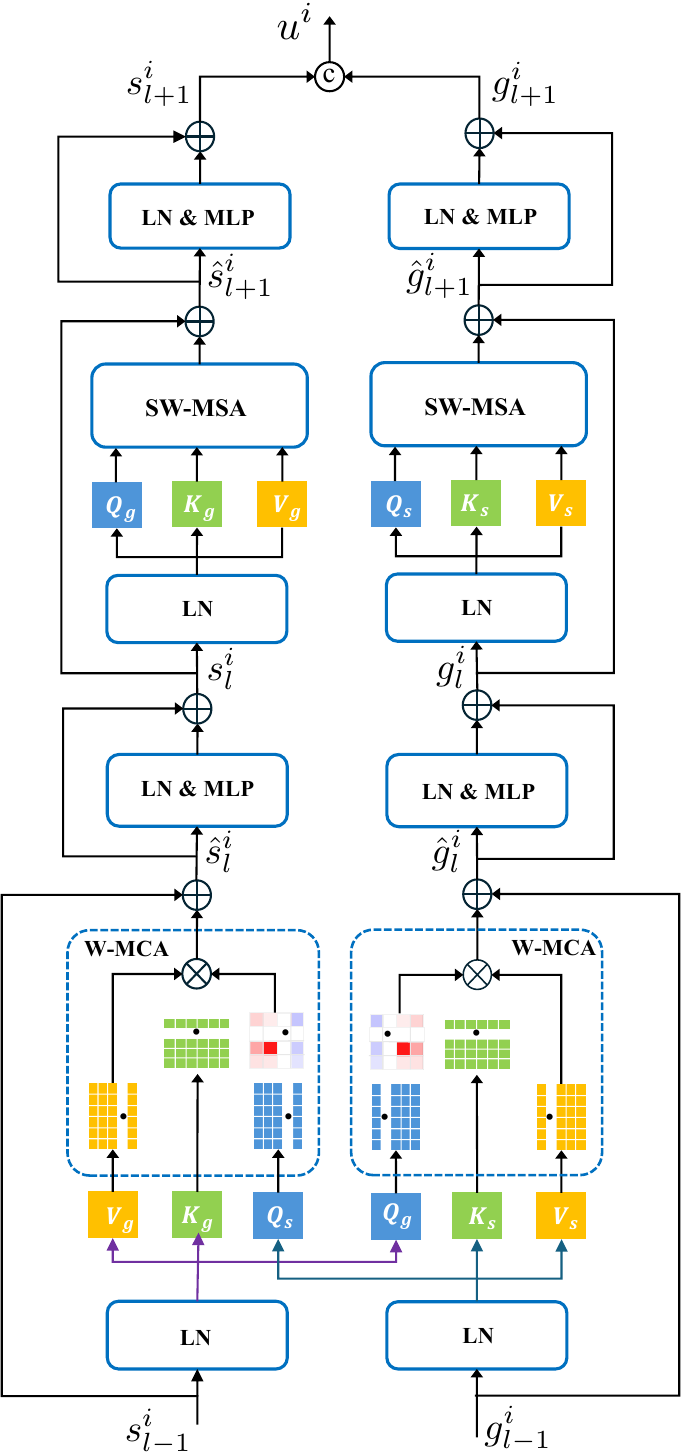}
\centering
\caption{The architecture of the proposed latent and global feature cross-attention (LGCA) module.} 	\label{fig3}
\end{figure}

The LGCA module efficiently fused the latent and global features to fully capture the relationship between them. As shown in Fig. \ref{fig3}, the LGCA module consisted of left and right branches, which encoded latent and global features from the upper and lower streams of the dual-stream feature learning-based encoder, respectively. The configurations of the left and right branches were the same as those in the global feature extraction module, except for the W-MSA, which was replaced with window-based multi-head cross-attention (W-MCA) to enable latent and global feature interaction. Specifically, after the linear embedding operations were applied to $s_{l-1}^i$ and $g_{l-1}^i$, two corresponding groups of \textit{query}, \textit{key}, \textit{value} matrics, $(Q_s,K_s,V_s)$ and $(Q_g, K_g, V_g)$, were obtained. Then,  multi-head cross-attention was calculated by swapping the \textit{query}s $Q_s$ with $Q_g$ in both branches, through
\begin{equation}
\textbf{Cross-Attention}(Q_g,K_s,V_s)=\mathrm{softmax}(\frac{Q_gK_s^\top}{\sqrt{d}})V_s, 
\end{equation}
and
\begin{equation}
\textbf{Cross-Attention}(Q_s,K_g,V_g)=\mathrm{softmax}(\frac{Q_sK_g^\top}{\sqrt{d}})V_g.
\end{equation}
After performing the operations of adding, $LN(\cdot)$ and $MLP(\cdot)$ on the outputs of W-MCA modules, $\hat{s}_l^i$ and $\hat{g}_l^i$ were obtained, which could be expressed as:
\begin{equation}
\begin{aligned}
	\hat{\mathbf{s}}^i_l& =\text{W-MCA}(\mathrm{LN}(\mathbf{s}^i_{l-1}),\mathrm{LN}(\mathbf{g}^i_{l-1}))+\mathbf{s}^i_{l-1}, \\
	\hat{\mathbf{g}}^i_l& =\text{W-MCA}(\mathrm{LN}(\mathbf{g}^i_{l-1}),\mathrm{LN}(\mathbf{s}^i_{l-1}))+\mathbf{g}^i_{l-1}. \\
\end{aligned}
\end{equation}
Here $\text{W-MCA}(\cdot,\cdot)$ represented regular-window partitioning configuration-based multi-head cross-attention. $\mathbf{\hat{s}}^i_{l}$ and $\mathbf{\hat{g}}^i_{l}$ denoted the outputs of the left and right W-MCA at the $i$-th LGCA module, respectively. After applying the same calculations on $\mathbf{\hat{s}}^i_{l}$ and $\mathbf{\hat{g}}^i_{l}$ as the last three formulas of Eq.~(\ref{eq7}) followed by a concatenation operation, the final result of LGCA could be obtained.

\subsubsection{Upsampling and convolution-based decode}
The upsampling and convolution-based decoder consisted of sequential layers, each integrating upsampling and convolution with a uniform convolutional kernel size of $3 \times 3$. It processed the final LGCA module outputs using five upsampling and convolution operations, producing the final deformation field $\phi$ with the exact resolution as the original image. We also concatenated the feature maps from each layer of the decoder with those from the LGCA module to effectively utilize the learned features. Additionally, we adopted one convolutional layer on the original image pair to produce low-level features and concatenated them with the features generated from the decoder's penultimate layer to further enhance the network's representation capability. 


\subsection{Loss function}
\label{LossFunction}
The whole network was trained in an unsupervised-learning manner, and the loss function consisted of two components, one being an image-similarity term to quantify the dissimilarity between the warped moving image $M\circ\phi^{-1}$ and fixed image $F$, and the other being a regularizer to penalize local spatial variations in the deformation field $\phi$. The final objective function used for LDM-Morph training was constructed as
\begin{equation}
\label{eq12}
\mathcal{L}_{REG}=\min_{\phi}\mathcal{L}_{sim}(M\circ\phi^{-1},F)+\lambda\mathcal{L}_{smooth}(\phi),
\end{equation}
where $\mathcal{L}_{sim}(\cdot)$ was the image similarity term, $\mathcal{L}_{smooth}(\cdot)$ was the regularizer, and $\lambda$ was a weighting fact to balance these two components. 

Mean squared error (MSE) and normalized cross-correlation (NCC) were widely used as similarity components in the loss functions of DLR methods. 
However, dissimilarity measured in the original pixel space might not fully capture the correlation between image pairs, potentially leading to deformation folding and anatomical mismatching. To address this issue, we focused on MSE and proposed a hierarchical similarity metric that quantified image dissimilarity in both the original and latent spaces, leveraging the encoder trained in the LDM. The hierarchical similarity metric was calculated as
\begin{equation}
\label{eq13}
\mathcal{L}_{sim}(M\circ\phi^{-1},F)=\beta\mathcal{L}_{org}+(1-\beta)\mathcal{L}_{lat}, 
\end{equation}
where
\begin{equation}
\mathcal{L}_{org}=\frac{1}{|\Omega|}\sum_{\mathbf{p}\in\Omega}\beta\left[M\circ\phi^{-1}(\mathbf{p})-F(\mathbf{p})\right]^2,
\end{equation}
and
\begin{equation}
\mathcal{L}_{lat}=\frac{1}{|\Omega|}\sum_{\mathbf{p}\in\Omega}(1-\beta)\left[E\left(M\circ\phi^{-1}(\mathbf{p})\right)-E\left(F(\mathbf{p})\right)\right]^2.
\end{equation}
Here $E(\cdot)$ was the encoder of the trained LDM and $\beta$ was the weight balancing $\mathcal{L}_{org}$ and $\mathcal{L}_{lat}$. For the second term of Eq.~(\ref{eq12}), a diffusion-based regularizer was applied to the spatial gradients of displacement $\phi$
\begin{equation}
\mathcal{L}_{smooth}(\phi)=\sum_{\mathbf{p}\in\Omega}\lVert\nabla\phi(\mathbf{p})\rVert^2.
\end{equation}

\section{Experiments}
\label{section4}
\subsection{Datasets}
Four public datasets were utilized to evaluate the performance of the proposed method on 2D medical image registration. The first two datasets came from the large-scale CAMUS echocardiography database, which included 2D apical four-chamber (CAMUS-4CH) and apical two-chamber (CAMUS-2CH) images from 500 patients~\citep{leclerc2019deep}. For each image, the endocardium and epicardium of the left ventricle and the left atrium walls were manually delineated. Our experiments involved registering the end-diastolic images to their corresponding end-systolic images. Regarding preprocessing, all images were resized to the size of $112\times112$, then padded to the size of $128\times128$. Pixel values were normalized to $[0,1]$. Both CAMUS-2CH and CAMUS-4CH were partitioned into 400, 20, and 80 pairs for training, validation, and testing, respectively.

The third dataset, EchoNet-Dynamic, consisted of 10,024 apical four-chamber 2D grayscale echocardiographic videos, each corresponding to a unique individual~\citep{ouyang2020video}. The left ventricle was manually annotated in each frame. Here the registration task was to align the end-diastolic frames to their corresponding end-systolic frames. The image resizing and intensity normalization operations were the same as the CAMUS datasets mentioned above. This dataset was divided into 7,460, 1,288, and 1,276 pairs for training, validation, and testing, respectively. 

The fourth dataset came from the Automated Cardiac Diagnosis Challenge (ACDC)~\citep{bernard2018deep}, which included 150 cardiac MRI 3D images acquired from 150 patients with manual annotations of the left ventricle, right ventricle, and myocardium. Single-slice pairs were extracted from the end-diastolic and the corresponding end-systolic 3D images, yielding 150 data pairs. The image resizing and intensity normalization operations were the same as the CAMUS datasets mentioned above. The dataset was further partitioned into 95, 5, and 50 pairs for training, validation, and testing, respectively.

Additionally, to evaluate the generalizability of the proposed method, we performed an experiment to directly apply the network trained on the EchoNet-Dynamic dataset to the testing CAMUS-2CH and CAMUS-4CH datasets. 


\subsection{Implementation details}
The proposed method was implemented using PyTorch based on the NVIDIA A100 GPU. The architecture of the LDM was configured to be the same as in~\citep{rombach2022high}. Specifically, a vector quantized generative adversarial network (VQGAN)~\citep{esser2021taming} was employed in LDM to encode the original image to the latent space and vice versa. The VQGAN produced a single feature map, with the feature resolution reduced to one-quarter of the original image's resolution. For the denoising U-Net, the number of layers in each encoder block was set to 3 with a channel size of 64. Features from the first and the third layers were empirically extracted at the time of $t=1$ and embedded into the upper stream of the proposed model. The Adam optimizer was employed for network training with a learning rate of $1\times10^{-4}$ and a batch size of 1. For the loss function, $\lambda$ and $\beta$ were set to 0.01 and 0.6, respectively, for all the experiments. For all models,  the checkpoint with the highest DSC value on the validation datasets was selected for inference. 
\subsection{Evaluation metrics}
Two widely adopted criteria were utilized as the evaluation metrics, including the Dice Similarity Coefficient (DSC) and the percentage of points in the deformation field $\phi$ with non-positive Jacobian determinant $|J_{\phi}|\leq0$ $(\%)$. Given a warped moving segmentation map $M_s^{'}=M_s\circ{\phi^{-1}}$ and a target segmentation map $T_s$, DSC calculated the structure overlap as
\begin{equation}
DSC=\frac{2\times|M_s^{'}\cap{T_s|}}{|M_s^{'}|+|T_s|}.
\end{equation}
A higher DSC indicated a better registration accuracy. For the first three datasets, DSC values were calculated based on the left ventricle. For the ACDC dataset, the average DSC values across the left ventricle and the myocardium were evaluated. 

The Jacobian matrix $J_{\phi}(p)$ captured local properties of $\phi$ surrounding point $p$ and was given by
\begin{equation}
J_{\phi}(p)=\begin{pmatrix}\frac{\partial{\phi}_x(p)}{\partial p_x}&&\frac{\partial{\phi}_x(p)}{\partial p_y}\\\frac{\partial{\phi}_y(p)}{\partial p_x}&&\frac{\partial{\phi}_y(p)}{\partial p_y}\end{pmatrix}.
\end{equation}
A non-positive determinant value of Jacobian matrix $|J_{\phi}(p)|$ denoted a fold at $p$, indicating the topology destruction at that point. A lower $|J_{\phi}|\leq0$ $(\%)$ value indicated a better topology-preservation ability.

\subsection{Reference methods}
To evaluate the performance of the proposed method, we compared it against two widely used, well-established traditional algorithms, SyN~\citep{avants2008symmetric} and LDDMM~\citep{beg2005computing}, both of which are widely used optimization-based diffeomorphic registration techniques. In our experiments, the Insight Toolkit (ITK)-based implementation of SyN and a PyTorch-based version of LDDMM were utilized. NCC and MSE were set as their cost functions, respectively. SyN was configured with a multi-resolution cascading registration using 160, 80, and 40 iterations with the following command: "ANTS 3 -m CC[fixed,moving,1,2] -t SyN[0.25] -r Gauss[3,0.25] -o output -i 160x80x40 --continue-affine false". LDDMM employed five time-dependent velocity fields with 500 iterations.

Additionally, VoxelMorph~\citep{balakrishnan2019voxelmorph}, CycleMorph~\citep{kim2021cyclemorph}, TransMorph~\citep{chen2022transmorph}, TransMatch~\citep{chenz2023transmatch}, and DiffuseMorph~\citep{kim2022diffusemorph} were adopted as the comparative DLR methods. VoxelMorph and CycleMorph are popular CNN-based DLR methods, TransMorph and TransMatch represent state-of-the-art Transformer-based approaches, and DiffuseMorph is a leading diffusion models-based DLR method. We reimplemented the official code for these 5 models in a 2D version, maintaining all other configurations as their default settings. Specifically, MSE was utilized as the similarity metric in the loss function for VoxelMorph, TransMorph, TransMatch, and DiffuseMorph, while NCC was utilized as the similarity metric for CycleMorph. These settings yielded the best results across all datasets in our experiments.

\section{Results}
\label{section5}
\begin{table*}[!t]
\centering
\caption{Quantitative results on mean and standard deviations of DSC values and the percentage of pixels with non-positive Jacobian determinants ($\%$), as well as the average computational time (seconds) per registration on the CAMUS-2CH, CAMUS-4CH, ECHO, and ACDC datasets obtained from different methods.}
\label{table1}
\large
\renewcommand{\arraystretch}{1.3}
\resizebox{\textwidth}{!}{%
	\begin{tabular}{ccccccccccccc}
		\hline \hline
		\multirow{2}{*}{Method} & \multicolumn{3}{c}{CAMUS-2CH} & \multicolumn{3}{c}{CAMUS-4CH} & \multicolumn{3}{c}{ECHO} & \multicolumn{3}{c}{ACDC} \\ \cline{2-13} 
		& Avg. DSC & $|J_\phi| \leq 0$ (\%) & Time (s) & Avg. DSC &  $|J_\phi| \leq 0$ (\%) & Time (s) & Avg. DSC &  $|J_\phi| \leq 0$ (\%) & Time (s) & Avg. DSC &  $|J_\phi| \leq 0$ (\%) & Time (s) \\ \hline
		Initial                  & 0.764 (0.067) & -           & -     & 0.764 (0.071) & -           & -     & 0.738 (0.076) & -           & -     & 0.555 (0.207) & -           & -     \\
		SyN                     & 0.853 (0.064) & 1.460 (0.576) & 2.05  & 0.870 (0.064) & 1.649 (0.600) & 2.06  & 0.860 (0.066) & 1.190 (0.681) & 2.13  & 0.799 (0.152) & 0.677 (0.475) & 1.79  \\
		LDDMM                   &0.856(0.063) & 0.035 (0.201) & 6.35  & 0.859 (0.078) &  $<$ 0.01         & 6.39  & 0.840 (0.109) & 0.014 (0.189) & 6.52  & 0.798 (0.115) & 0.013 (0.057) & 6.31  \\ \hline
		VoxelMorph              & 0.852 (0.054) & 0.542 (0.499) & 0.03  & 0.858 (0.061) & 0.607 (0.452) & 0.03  & 0.857 (0.061) & 0.802 (0.659) & 0.01  & 0.830 (0.102) & 0.375 (0.492) & 0.04  \\
		CycleMorph              & 0.824 (0.059) & 0.945 (0.586) & 0.03  & 0.842 (0.061) & 0.537 (0.381) & 0.03  & 0.850 (0.061) & 1.365 (0.898) & 0.01  & 0.784 (0.111) & 0.302 (0.323) & 0.05  \\
		DiffuseMorph            & 0.852 (0.052) & 0.437 (0.417) & 0.08  & 0.852 (0.062) & 0.728 (0.464) & 0.05  & 0.838 (0.071) & \textbf{0.302 (0.377)} & 0.05  & 0.664 (0.137) & 0.390 (0.419) & 0.07  \\
		TransMorph              & 0.871 (0.051) & 0.696 (0.450) & 0.04  & 0.876 (0.057) & 0.842 (0.476) & 0.04  & 0.863 (0.058) & 1.753 (1.033) & 0.02  & 0.842 (0.076) & 0.319 (0.433) & 0.05  \\
		TransMatch              & 0.871 (0.052) & 0.523 (0.393) & 0.06  & 0.875 (0.059) & 0.777 (0.421) & 0.06  & 0.874 (0.058) & 1.331 (0.892) & 0.05  & 0.841 (0.076) & 0.344 (0.466) & 0.07  \\
		Ours              & \textbf{0.882 (0.049)} & \textbf{0.239 (0.263)} & 0.11  & \textbf{0.889 (0.055)} & \textbf{0.441 (0.351)} & 0.11  & \textbf{0.883 (0.056)} & 0.400 (0.419) & 0.10  & \textbf{0.850 (0.053)} & \textbf{0.157 (0.255)} & 0.12  \\ \hline \hline
	\end{tabular}%
}
\end{table*}

\begin{figure*}[!t]
\setlength{\abovecaptionskip}{-6pt}
\setlength{\belowcaptionskip}{0pt}
\includegraphics[width=16cm]{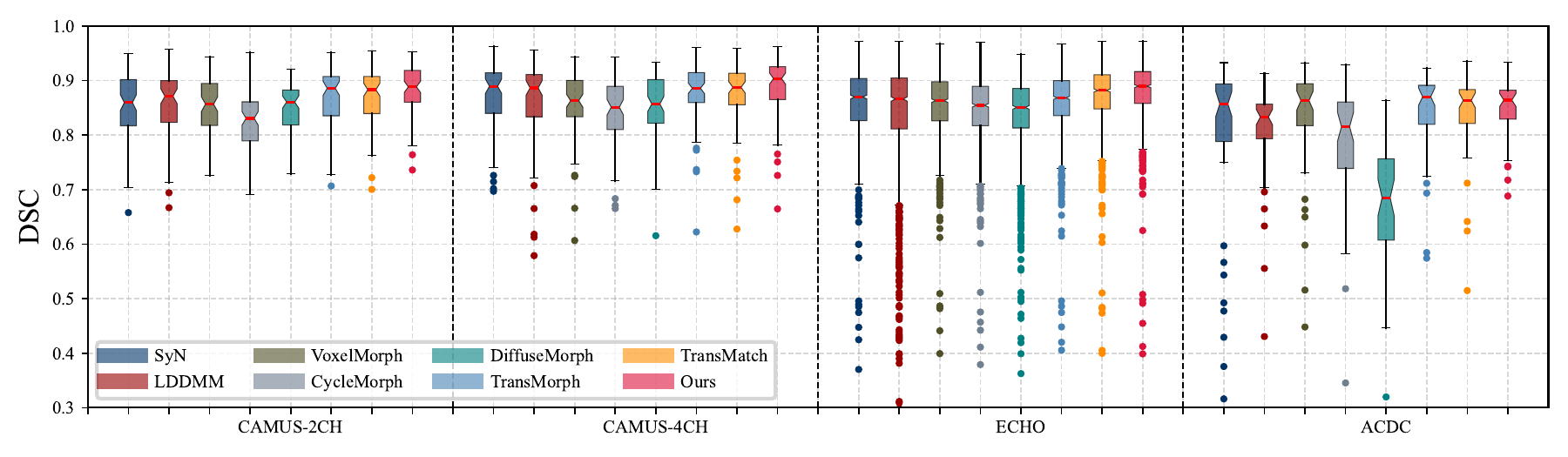}
\centering
\caption{Boxplot of DSC values for registration results on the CAMUS-2CH, CAMUS-4CH, ECHO, and ACDC datasets obtained from different methods. The red line denotes the median. The upper and lower boundaries of each rectangular box represent the upper and lower quartiles, respectively. The whiskers indicate the range of the data.} \label{fig4}
\end{figure*}

\subsection{Comparisons with state-of-the-art approaches}

The quantitative results of different methods, including the mean and standard deviations of DSC, the percentage of voxels with non-positive Jacobian determinants, and the average computational time per registration (in seconds) across four datasets, are presented in Table~\ref{table1}. The proposed LDM-Morph demonstrated superior performance in average DSC values compared to the other seven approaches evaluated. Specifically, it achieved average DSC values of 0.882 for CAMUS-2CH, 0.889 for CAMUS-4CH, 0.883 for ECHO, and 0.850 for ACDC. Compared to traditional algorithms (SyN and LDDMM), the proposed LDM-Morph exceeded by at least 1.9\% on the first three datasets and by 5\% on the ACDC dataset. Compared to CNN-based DLR methods (VoxelMorph and CycleMorph), LDM-Morph consistently outperformed them with margins of at least 2\% across all four datasets. In comparison to the diffusion models-based registration method (DiffuseMorph), LDM-Morph had DSC values at least 3\% higher across all datasets. With reference to Transformer-based approaches (TransMorph and TransMatch), LDM-Morph had approximately 1\% higher average DSC values. Additionally, LDM-Morph achieved the lowest standard deviation in DSC values, highlighting its robustness and consistency across different datasets.

\begin{figure*}[!t]
\centering
\includegraphics[width=16cm]{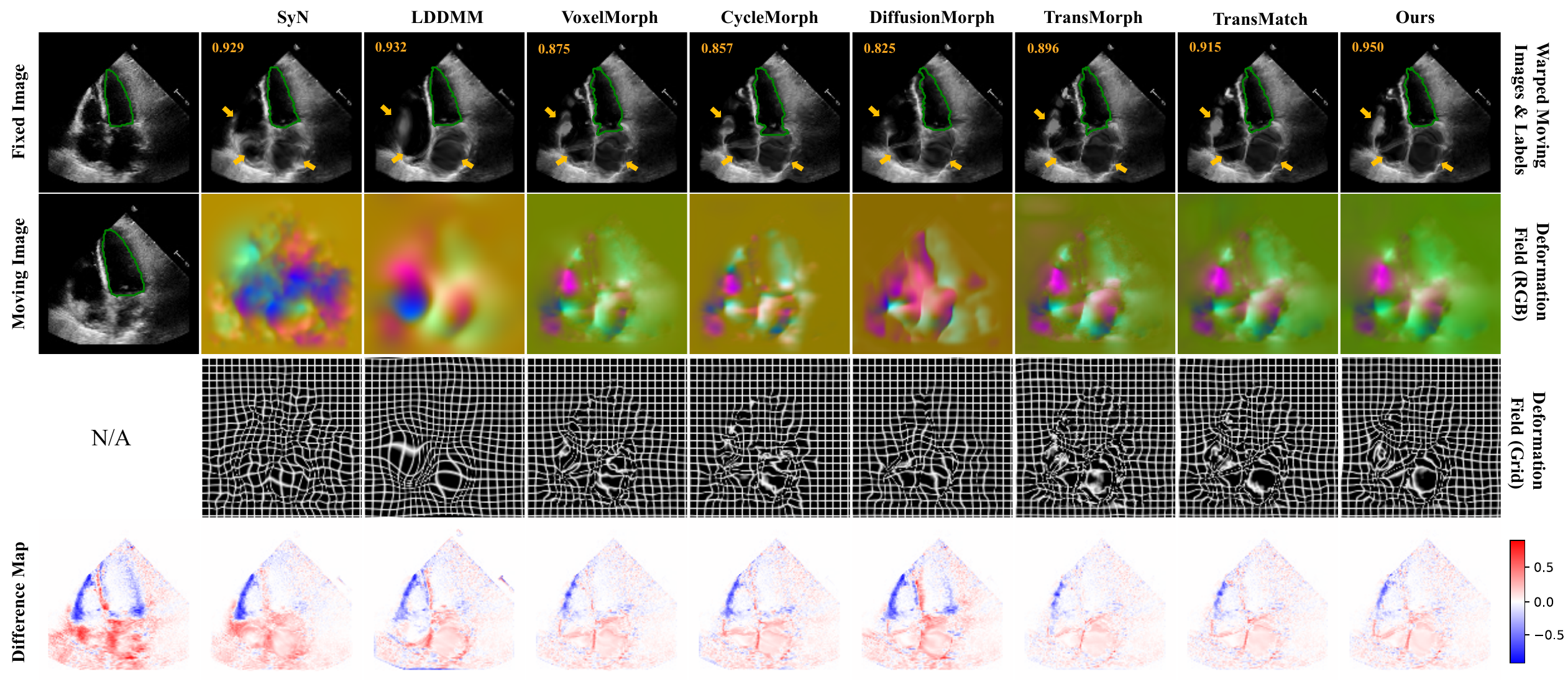}
\caption{Visual comparison of registration methods on the ECHO dataset. The first column displays: the fixed image with segmentation overlay, the moving image with segmentation overlay, and the difference between the moving and fixed images. Subsequent columns present results from SyN, LDDMM, VoxelMorph, CycleMorph, DiffusionMorph, TransMorph, TransMatch, and the proposed LDM-Morph. The DSC values are shown in the upper left corner. From top to bottom: the warped moving images with overlaid warped segmentation, estimated deformations (in RGB), estimated deformations (as grids), and the differences between the warped moving images and the fixed image.} \label{fig5} 
\end{figure*}

Figure~\ref{fig4} shows the distribution of DSC values for the eight methods across four different datasets. In the first three datasets, LDM-Morph achieved a higher median DSC and a more compact distribution than the other seven methods. 
Although the median DSC value of LDM-Morph was slightly lower than that of TransMorph on the ACDC dataset, the distribution of DSC values of LDM-Morph was more centered. These observations collectively demonstrated the robustness of LDM-Morph. A representative registration sample from the ECHO dataset is presented in Figure~\ref{fig5}. The warped moving image generated by LDM-Morph was visually closest to the target image and had the highest DSC value. Additionally, despite SyN and LDDMM exhibiting higher DSC values compared to the other five DLR methods, the differences between the warped and fixed images were more pronounced in these methods than in VoxelMorph, TransMorph, TransMatch, and the proposed LDM-Morph, particularly in the regions indicated by the arrows. Additional qualitative comparison of various registration methods on the representative image pairs from the other three datasets can be found in~\ref{app1}.

Regarding topology preservation, LDM-Morph outperformed four other DLR methods across three datasets, except for the ECHO dataset, where a slightly higher number of pixels with non-positive Jacobian determinants was observed compared to DiffuseMorph. Compared to the traditional LDDMM algorithm, similar to other DLR methods, LDM-Morph had a much higher percentage of pixels with $|J_{\phi}|\leq 0$. 
This was due to LDDMM optimizing a series of time-dependent velocity fields during the registration process and using a Laplacian differential operator for regularization~\citep{beg2005computing}. While these techniques penalized local inconsistencies, they also lowered the DSC values for LDDMM. In terms of computational time, while the proposed LDM-Morph exhibited a marginally higher average than other DLR methods, it was still significantly faster than the traditional algorithms.

\subsection{Evaluation on the generalization performance}

To evaluate the generalizability of the DLR methods, six DLR models trained on the ECHO dataset were applied directly to the CAMUS-2CH and CAMUS-4CH datasets. Quantitative results in Table~\ref{table2} show that all methods exhibited good generalizability when evaluated on datasets different from their training data.  Clearly, the proposed LDM-Morph demonstrated superior generalization performance, achieving the highest DSC values and the lowest percentage of pixels with $|J_\phi| \leq 0$ in the deformation fields across both datasets. In contrast, the DiffusionMorph consistently showed lower performance than the other methods, with a notable drop in registration accuracy when applied across different datasets, compared to when it was trained and tested on the same dataset. This decline stemmed from DiffusionMorph's heavy reliance on spatial information derived from the original pixel space using DDPM, an intensity-sensitive approach that limited its generalizability across varying datasets. In comparison, LDM-Morph mitigated this issue by leveraging LDM to perform the diffusion process in a latent space, focusing on high-level features. This strategy improved the generalizability of the model across diverse datasets, leading to enhanced overall registration performance.

\begin{table}[!t]
\centering
\setlength{\belowcaptionskip}{-0.cm}
\caption{Quantitative results on mean and standard deviations of DSC values and the percentage of voxels with non-positive Jacobian determinants ($\%$) by applying the different models trained using the ECHO dataset on the CAMUS-2CH and CAMUS-4CH datasets.}
\label{table2}
\renewcommand{\arraystretch}{1.3}
\resizebox{\columnwidth}{!}{%
	\scriptsize
	\begin{tabular}{ccccc} \\ \hline  \hline
		\multirow{2}{*}{Method} & \multicolumn{2}{c}{CAMUS-2CH}   & \multicolumn{2}{c}{CAMUS-4CH}   \\ \cline{2-5} 
		& Avg.   DSC      & $|J_\phi| \leq 0$ (\%)  & Avg. DSC        & $|J_\phi| \leq 0$ (\%)   \\ \hline
		Initial                  & 0.764   (0.067) & -           & 0.764   (0.071) & -          \\
		SyN                     & 0.853 (0.064)   & 1.460 (0.576) & 0.870 (0.064)   & 1.649 (0.600) \\
		LDDMM                   & 0.856 (0.063)   & 0.035 (0.201) & 0.859 (0.078)   & $<$ 0.01         \\ \hline
		VoxelMorph              & 0.842 (0.060)   & 0.233 (0.198) & 0.851 (0.062)   & 0.300 (0.193) \\
		CycleMorph              & 0.841 (0.057)   & 0.590 (0.337) & 0.851 (0.062)   & 0.805 (0.428) \\
		DiffuseMorph            & 0.800 (0.060)   & 0.466 (0.314) & 0.806 (0.071)   & 0.385 (0.273) \\
		TransMorph              & 0.850 (0.056)   & 1.059 (0.570) & 0.861 (0.059)   & 1.401 (0.634) \\
		TransMatch              & 0.872 (0.058)   & 0.780 (0.562) & 0.877 (0.060)   & 1.066 (0.607) \\
		Ours                    & \textbf{0.877 (0.053)}   & \textbf{0.166 (0.278)} & \textbf{0.889 (0.054)}   & \textbf{0.178 (0.185)} \\ \hline \hline
	\end{tabular}%
	}
\end{table}

\begin{figure*}[!t]
\includegraphics[width=16cm]{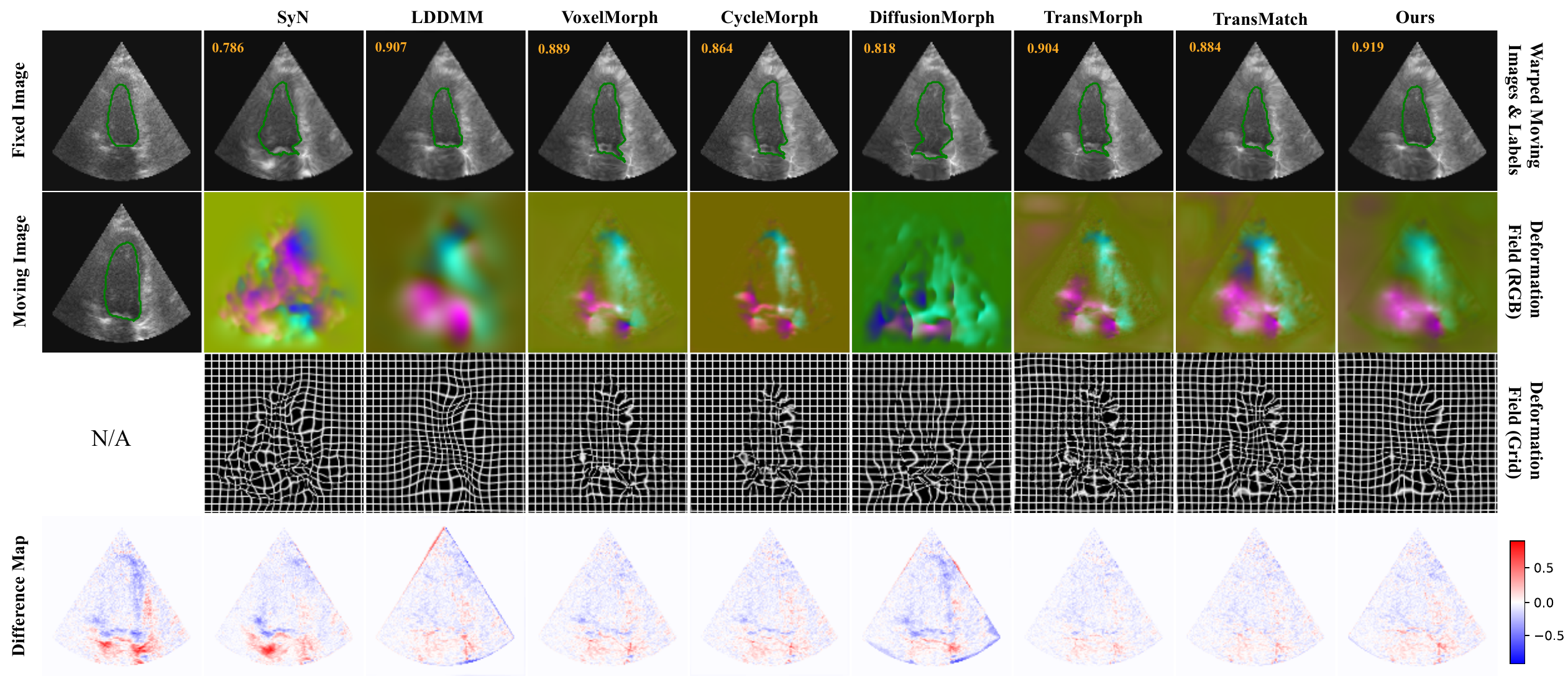}
\centering
\caption{Visual comparison of different registration methods on the CAMUS-2CH dataset. The first column displays: the fixed image with segmentation overlay, the moving image with segmentation overlay, and the difference between the moving and fixed images. Subsequent columns present results from SyN, LDDMM, VoxelMorph, CycleMorph, DiffusionMorph, TransMorph, TransMatch, and the proposed LDM-Morph. The DSC values are shown in the upper left corner. From top to bottom: the warped moving images with overlaid warped segmentation, estimated deformations (in RGB), estimated deformations (as grids), and the differences between the warped moving images and the fixed image.} \label{fig6}
\end{figure*}

Moreover, the Transformer-based methods exhibited better generalizability than the CNN-based methods. The multi-head self- and cross-attention mechanisms in TransMatch secured it a second place among all comparative approaches regarding average DSC values. However, the numbers of pixels with $|J_\phi|\leq0$ in TransMatch for both datasets were much higher than those in Table~\ref{table1}, wherein these values of LDM-Morph were lower, further demonstrating the superior performance of LDM-Morph in registration, accuracy and topology preservation, even in cross-dataset registration scenarios. Conventional registration algorithms (SyN and LDDMM) delivered consistent performance across both datasets, as they were optimized individually for each image pair. Figure~\ref{fig6} provides a visual comparison of various registration methods under the generalization experimental settings. This comparison demonstrates that LDM-Morph produced registration results that were visually closest to the fixed image, achieving the highest DSC. Additionally, LDM-Morph generated a smoother deformation field than other DLR methods, underscoring its superior generalization performance.

\begin{figure*}[!t]
\includegraphics[width=16cm]{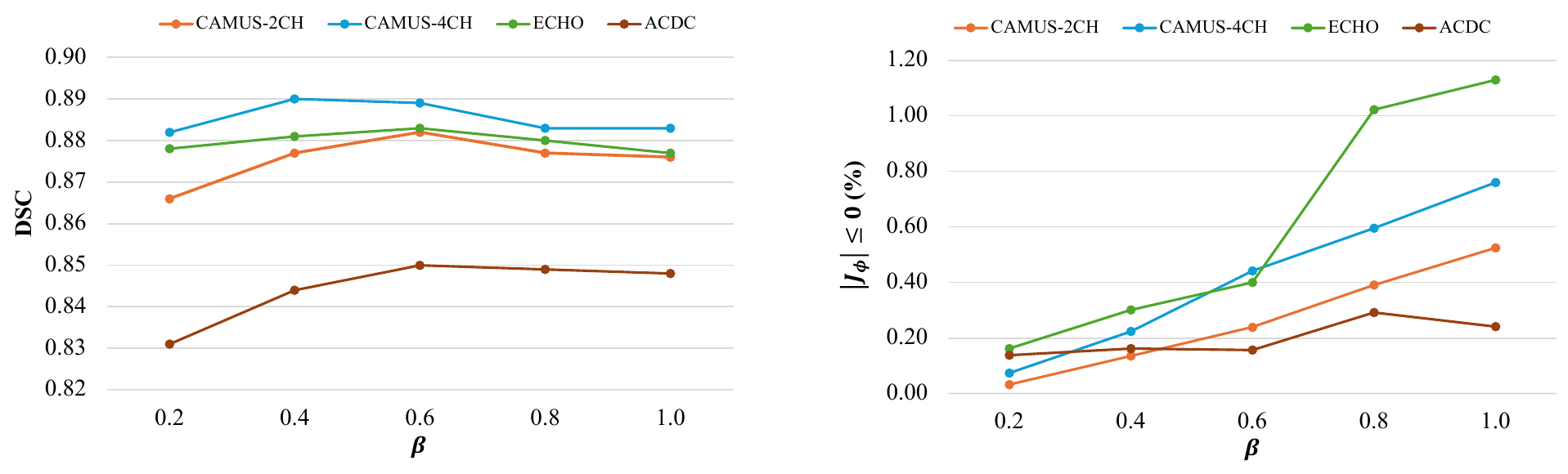}
\centering
\caption{Average DSC values and corresponding average number of pixels with $|J_\phi|\leq0$ of four testing datasets for our method with varied parameter $\beta$ in the hierarchical similarity metric.} \label{fig7}
\end{figure*}

\subsection{Effects of the hierarchical similarity metric}
\label{sec443}
To examine the impact of the hierarchical similarity metric in the loss function on the performance of LDM-Morph, we conducted a comparative study by varying the value of $\beta$ in Eq.~(\ref{eq13}) across all four datasets. Fig.~\ref{fig7} displays the average DSC values alongside the corresponding percentage of the number of pixels with $|J_\phi|\leq0$, evaluated on four different datasets based on different $\beta$ values. The left panel shows that the average DSC values increased for all four datasets as $\beta$ increased. The highest DSC values were achieved at $\beta=0.6$ for the CAMUS-2CH, ECHO, and ACDC datasets, while for the CAMUS-4CH dataset, the highest DSC value was reached at $\beta=0.4$. Beyond these points, the DSC values gradually declined. 
The right panel of Fig.~\ref{fig7} shows that, except for the ACDC dataset, the $|J_\phi|\leq0$ (\%) values consistently increase as the $\beta$ value increases from 0.2 to 1.0. In particular, for the ECHO dataset, the $|J_\phi| \leq 0$ (\%) at $\beta = 0.8$ is more than twice that at $\beta = 0.6$. Although the $|J_\phi| \leq 0$ (\%) at $\beta = 1.0$ is lower than that at $\beta = 0.8$ for the ACDC dataset, it remains higher than all the $|J_\phi| \leq 0$ (\%) values for $\beta < 0.8$.  These results suggest that the proposed hierarchical similarity metric, particularly the latent space dissimilarity term $\mathcal{L}_{lat}$, is crucial in enhancing topological preservation and improving registration accuracy.

However, the increase in foldings within the deformation fields did not show an increase in average DSC values after $\beta=0.6$ (see right panel of Fig.~\ref{fig7}). The underlying reason was that at lower values of $\beta$, the MSE emphasized dissimilarities in the latent space, where high-level semantic information prevailed, resulting in smoother displacement fields but lower DSC values. In contrast, as $\beta$ increased, the focus shifted toward lower-level features, such as pixel values, which benefited registration accuracy. However, these lower-level features did not fully capture the anatomical information correlations within the image pair, ultimately leading to a decline in DSC values and more topology disruptions. Such findings demonstrated the effectiveness of the proposed hierarchical similarity metric of the loss function in enhancing registration performance. By carefully balancing the contributions of high-level semantic information and low-level pixel features through the appropriate selection of $\beta$, LDM-Morph successfully maintained smooth deformation fields while optimizing registration accuracy.

\subsection{Ablation studies}
A comprehensive evaluation was conducted to assess the efficacy of the proposed modules within LDM-Morph, which included the LDM-based latent feature extraction (LDM-FE) module, the Latent and Global Feature-based Cross-Attention (LGCA) module, and the similarity loss function term $\mathcal{L}_{lat}$. The LDM-FE module leveraged features extracted from a denoising U-Net, enhancing the feature diversity and improving the image alignment. The LGCA module employed a multi-head cross-attention mechanism to facilitate interactions between hierarchical information, effectively capturing the relationships between latent and global features during deformation. The term $\mathcal{L}_{lat}$ in the loss function was designed to assess alignment at the semantic level, ensuring that image pairs encoded by the LDM encoder retained detailed anatomical structure information and preserved the topology more effectively. The models were trained and evaluated separately for each dataset.

\begin{table*}[t]
\centering	
\caption{Quantitative results on mean and standard deviations of the DSC values and the percentage of voxels with non-positive Jacobian determinants ($\%$) of ablation study on different modules.}
\label{table3}
\renewcommand{\arraystretch}{1.3}
\resizebox{\textwidth}{!}{%
	\begin{tabular}{ccccccccccc} \\ \hline \hline
		\multirow{2}{*}{LDM-FE} & \multirow{2}{*}{LGCA} & \multirow{2}{*}{$\mathcal{L}_{lat}$} & \multicolumn{2}{c}{CAMUS-2CH}   & \multicolumn{2}{c}{CAMUS-4CH} & \multicolumn{2}{c}{ECHO}      & \multicolumn{2}{c}{ACDC}      \\ \cline{4-11} 
		&                       &                                    & Avg. DSC             & $|J_\phi| \leq 0$ (\%)  & Avg. DSC           & $|J_\phi| \leq 0$ (\%)  & Avg. DSC           & $|J_\phi| \leq 0$ (\%)  & Avg. DSC           & $|J_\phi| \leq 0$ (\%)  \\ \hline
		&                       &                                    & 0.872   (0.051) & 0.653 (0.516) & 0.873 (0.058) & 0.641 (0.408) & 0.872 (0.058) & 1.320 (0.910) & 0.833 (0.087) & 0.326 (0.416) \\
		$\checkmark$ 	&    &    & 0.874   (0.048) & 0.471(0.373)  & 0.881 (0.056) & 0.786 (0.483) & 0.874 (0.057) & 1.332 (0.926) & 0.845 (0.060) & 0.240 (0.311) \\
		$\checkmark$ 	&   $\checkmark$  &     & 0.876   (0.050) & 0.525 (0.438) & 0.883 (0.055) & 0.759 (0.491) & 0.877 (0.057) & 1.129 (0.914) & 0.848 (0.056) & 0.241 (0.334) \\
		$\checkmark$	&  $\checkmark$  &  $\checkmark$ & \textbf{0.882 (0.049)} & \textbf{0.239 (0.263)} & \textbf{0.889 (0.055)} & \textbf{0.441 (0.351)} & \textbf{0.883 (0.056)} & \textbf{0.400 (0.419)} & \textbf{0.850 (0.053)} & \textbf{0.157 (0.255)} \\ \hline \hline
	\end{tabular}%
}
\end{table*}

Table~\ref{table3} presents the quantitative results obtained by progressively incorporating the LDM-FE module, LGCA module, and the similarity loss function term $\mathcal{L}_{lat}$ into the baseline model, which initially included only the global feature learning branch, i.e., the lower stream (see Fig.~\ref{fig1}). Integrating all the proposed modules yielded the best performance. Notably, as each module was progressively added to the baseline model, the average DSC values consistently improved, highlighting the positive impact of the proposed modules. Although the combination of the LDM-FE and LGCA modules improved registration accuracy compared to the LDM-FE module alone, the $|J\phi| \leq 0$ (\%) values remained similar across all datasets. However, incorporating $\mathcal{L}_{lat}$ significantly reduced the percentage of pixels with $|J\phi| \leq 0$. As explained in Sec.~\ref{sec443}, this improvement was due to models trained solely with pixel-level similarity loss often struggling to maintain anatomical topology consistency. The introduction of $\mathcal{L}_{lat}$ mitigated this issue by enabling more effective semantic-level alignment, which reduced negative Jacobian determinants and enhanced overall registration accuracy.

\section{Discussion}
\label{section6}
Among traditional algorithms, SyN and LDDMM demonstrated strong performance in both accuracy and topology preservation. Therefore, we compared the proposed LDM-Morph with these two algorithms. LDM-Morph outperformed both in registration precision and showed promising topology preservation across all datasets, especially when compared to SyN. VoxelMorph, a well-known deformable image registration framework, employed a U-Net architecture to predict deformation fields. However, the small kernel size of its convolutional operations limited its performance~\citep{jia2022u}. Transformer-based models, such as TransMorph and TransMatch, addressed this limitation by incorporating the multi-head self-attention mechanism, resulting in improved performance in our experiments compared to CNN-based models like VoxelMorph and CycleMorph. Despite these improvements, Transformer-based models might produce more topology destruction than CNN-based models (see Tables~\ref{table1} and~\ref{table2}). The proposed LDM-Morph overcame these challenges by leveraging both latent and global features, achieving superior registration accuracy and better topology preservation.  

Our experimental results indicated that leveraging latent features from LDM could significantly enhance registration performance in medical image alignment.  Unlike existing CNN- and Transformer-based DLR methods, which focused on architecture innovations or designing multi-resolution cascading paradigms, our proposed LDM-Morph framework integrated features derived from the LDM directly into the network. This approach was proven highly effective across two distinct cardiac imaging modalities. Additionally, the integration of the proposed LGCA module strengthened the interaction between latent features from LDM and global features generated by the Transformer. This enhancement facilitated a better alignment of anatomical structures between the moving and fixed images during the deformation process, significantly improving registration accuracy, as demonstrated in our ablation studies (see Table~\ref{table3}). 

One contribution of this work is the introduction of a hierarchical similarity metric in the loss function. Unlike conventional DLR methods that used MSE or NCC evaluated merely on the original image space, our approach assessed dissimilarity between the warped moving image and the fixed image at two feature levels: low-level intensity features from the pixel space and high-level anatomical structure information from the latent space. By incorporating both intensity differences and semantic information into the deformation process, our method ensured high registration precision while maintaining a smooth deformation field. While latent feature extraction using LDM increased the computational load of the proposed registration model, the additional time—0.05 seconds per image pair compared to conventional Transformer-based models—is a small trade-off for the performance gains achieved.

One potential limitation of this work is that the proposed model was not constructed in an end-to-end manner. Integrating the LDM directly into the registration network, rather than training the LDM and registration network separately with their respective loss functions, can enable the model to learn more related features, potentially enhancing its performance. This is an area for future exploration. Additionally, evaluating the registration performance of the proposed LDM-Morph framework on other publicly available datasets from different modalities can further validate its capability and deserves further investigation. Finally, further extending the proposed LDM-Morph model to the 3D version and adapting it for general 3D medical image registration tasks is a promising direction and one of our future works.
\section{Conclusion}
\label{section7}
In this work, we proposed a novel unsupervised deformable registration framework, LDM-Morph, which estimated the optimal deformations by employing LDM as the feature extractor. Additionally,  an LGCA module was proposed to integrate latent and global information, and a hierarchical similarity metric was utilized to leverage information from multiple spaces.  Experimental results based on four public datasets show that the proposed LDM-Morph achieved state-of-the-art performance compared to other reference methods.

\section{Acknowledgments}
This work was supported by NIH grants R01EB034692 and R01AG078250.

\appendix
\onecolumn
\section{Additional results for 2D cardiac image registration}
\label{app1}
\begin{figure*}[h]
\includegraphics[width=17cm]{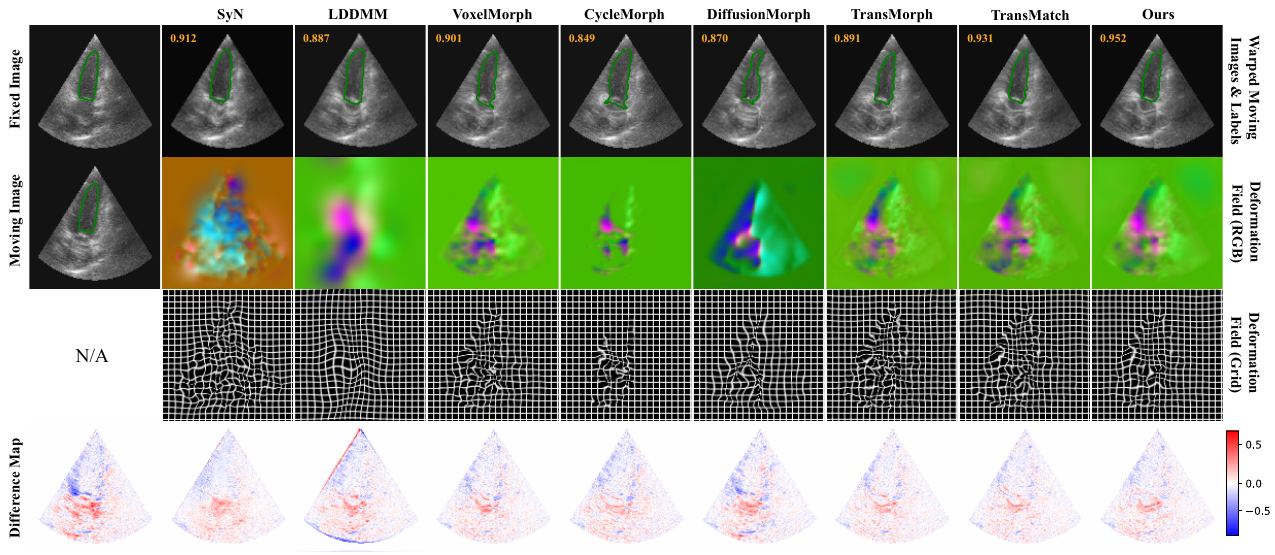}
\centering
\caption{ Additional qualitative comparison of various registration methods on the representative image pair from the CAMUS-2CH dataset.} \label{figa1}
\end{figure*}

\hspace{2cm}

\begin{figure*}[h]
\includegraphics[width=17cm]{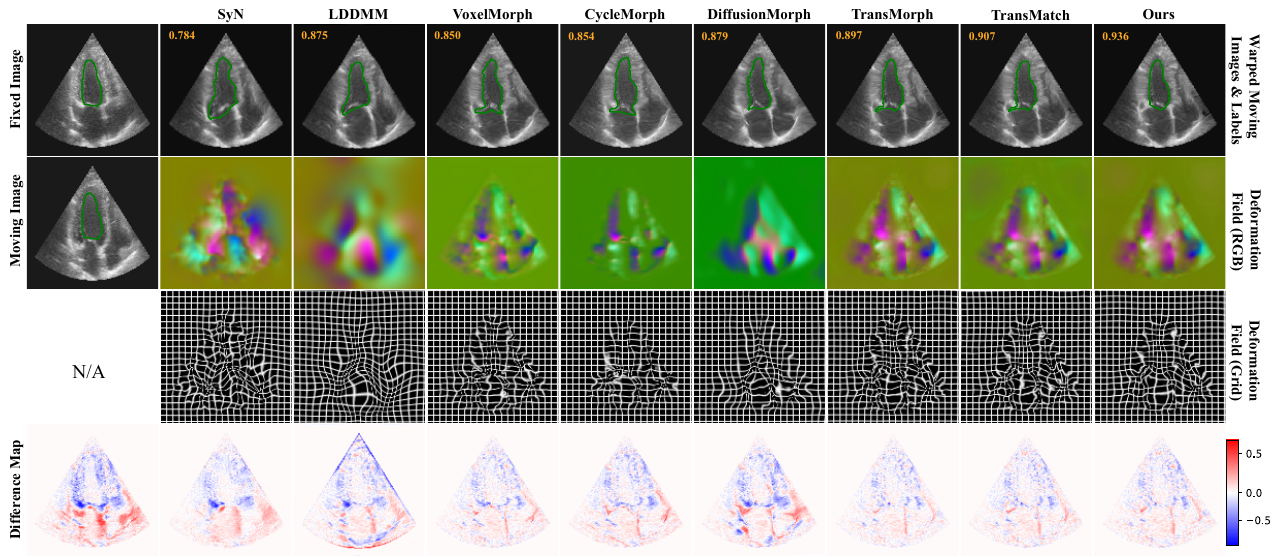}
\centering
\caption{ Additional qualitative comparison of various registration methods on the representative image pair from the CAMUS-4CH dataset.} \label{figa2}
\end{figure*}
\clearpage
\begin{figure*}[h]
\setlength{\belowcaptionskip}{-1cm} 
\includegraphics[width=17cm]{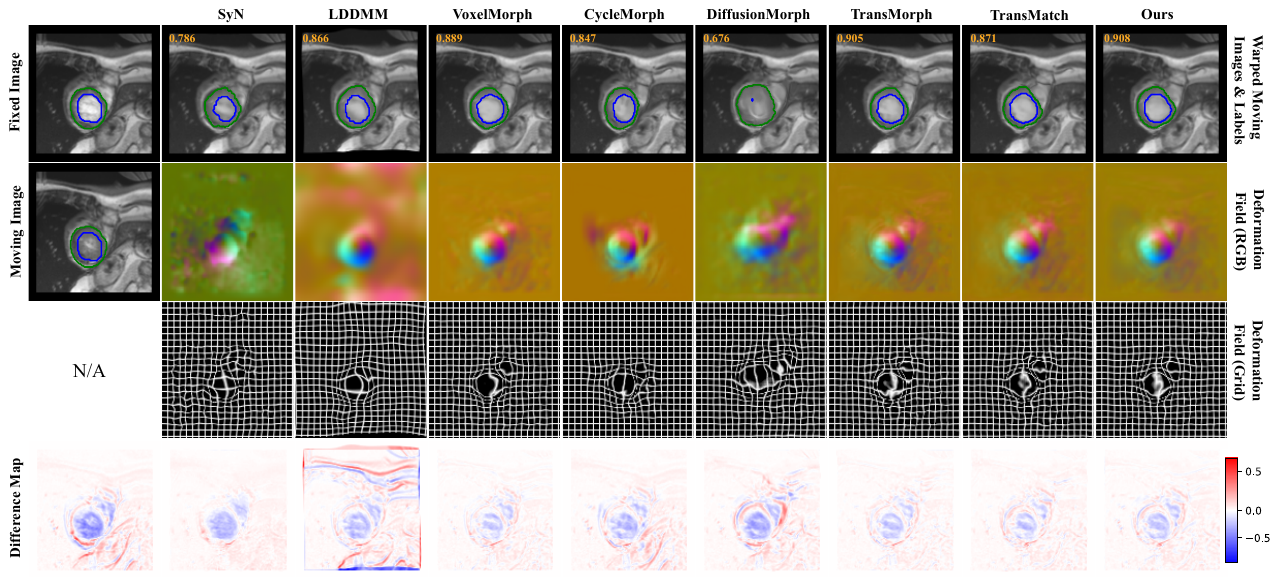}
\centering
\caption{Additional qualitative comparison of various registration methods on a representative image pair from the ACDC dataset.} \label{figa3}
\end{figure*}

\clearpage
\twocolumn

\bibliographystyle{elsarticle-harv}\biboptions{authoryear} 
\bibliography{refs}
%






\end{document}